\documentclass[sigconf]{acmart}

\usepackage{balance}

\usepackage{booktabs} 
\usepackage{multirow}
\usepackage{enumitem}

\usepackage[normalem]{ulem} 
\usepackage{xcolor}

\usepackage[linesnumbered]{algorithm2e}

\newcommand{\eg}{\textit{e.g.}}
\newcommand{\ie}{\textit{i.e.}}
\newcommand{\etal}{\textit{et al.}}
\newcommand{\systemname}{ProSeNet}

\newcommand{\systemurl}{https://github.com/myaooo/ProSeNet}

\def\mathbi#1{\textbf{\em #1}}

\DeclareMathOperator*{\argmin}{arg\,min}

\graphicspath{{figs/}{pictures/}{images/}{}{./}} 

\def\BibTeX{{\rm B\kern-.05em{\sc i\kern-.025em b}\kern-.08emT\kern-.1667em\lower.7ex\hbox{E}\kern-.125emX}}
    
\copyrightyear{2019} 
\acmYear{2019} 
\setcopyright{acmlicensed}
\acmConference[KDD '19]{The 25th ACM SIGKDD Conference on Knowledge Discovery and Data Mining}{August 4--8, 2019}{Anchorage, AK, USA}
\acmBooktitle{The 25th ACM SIGKDD Conference on Knowledge Discovery and Data Mining (KDD '19), August 4--8, 2019, Anchorage, AK, USA}
\acmPrice{15.00}
\acmDOI{10.1145/3292500.3330908}
\acmISBN{978-1-4503-6201-6/19/08}

\begin{document}
\title{Interpretable and Steerable Sequence Learning via Prototypes}


\author{Yao Ming}
\authornote{This work was done during his internship at Bosch Research North America.}
\affiliation{%
  \institution{Hong Kong University of Science and Technology}
}
\email{ymingaa@ust.hk}

\author{Panpan Xu}
\affiliation{%
  \institution{Bosch Research North America}
}
\email{panpan.xu@us.bosch.com}

\author{Huamin Qu}
\affiliation{%
  \institution{Hong Kong University of Science and Technology}
}
\email{huamin@cse.ust.hk}

\author{Liu Ren}
\affiliation{%
  \institution{Bosch Research North America}
}
\email{liu.ren@us.bosch.com}

\renewcommand{\shortauthors}{Yao Ming et al.}

\begin{abstract}
  One of the major challenges in machine learning nowadays is to provide predictions with not only high accuracy but also user-friendly explanations. Although in recent years we have witnessed increasingly popular use of deep neural networks for sequence modeling, it is still challenging to explain the rationales behind the model outputs, which is essential for building trust and supporting the domain experts to validate, critique and refine the model.\looseness=-1

  We propose \systemname, an interpretable and steerable deep sequence model with natural explanations derived from case-based reasoning. The prediction is obtained by comparing the inputs to a few \textit{prototypes}, which are exemplar cases in the problem domain. For better interpretability, we define several criteria for constructing the prototypes, including \textit{simplicity}, \textit{diversity}, and \textit{sparsity} and propose the learning objective and the optimization procedure. \systemname~also provides a \textit{user-friendly} approach to \textit{model steering}: domain experts without any knowledge on the underlying model or parameters can easily incorporate their intuition and experience by manually refining the prototypes. \looseness=-1
  
  We conduct experiments on a wide range of real-world applications, including predictive diagnostics for automobiles, ECG, and protein sequence classification and sentiment analysis on texts. The result shows that \systemname~can achieve accuracy on par with state-of-the-art deep learning models. We also evaluate the interpretability of the results with concrete case studies. Finally, through user study on Amazon Mechanical Turk (MTurk), we demonstrate that the model selects high-quality prototypes which align well with human knowledge and can be interactively refined for better interpretability without loss of performance.\looseness=-1
\end{abstract}

%
%

\ccsdesc[300]{Computing methodologies~Neural networks}
\ccsdesc[300]{Computing methodologies~Instance-based learning}


\keywords{Sequence learning; Deep Neural Network; Interpretability}

\maketitle

%
%

\section{Introduction}

Event sequence data is becoming pervasive in a variety of domains, \eg, electronic health records (EHR) in health care \cite{johnson2016mimic, choi2016rnn}, click streams in software applications and vehicle fault logs in automobiles.
In general, an event sequence is a series of temporally ordered events. 
With the advances of machine learning, especially deep learning, we have seen a growing trend of research that applies sequence learning to assist decision-making in these domains. For example, by modeling fault sequences collected from vehicle fleets, we can predict errors that are likely to occur in the future, and thus enable predictive maintenance for car manufacturers and repair workshops, which eventually could improve customer experience and reduce warranty costs \cite{chen2018sequence}. \looseness=-1

The most widely adopted method for modeling sequential data nowadays is Recurrent Neural Networks (RNNs) and its variants, such as Long Short-Term Memory networks (LSTMs). RNNs have achieved remarkable performance in various sequence modeling applications, \eg, document/text classification \cite{tang15sentiment}, machine translation \cite{sutskever14seq2seq} and speech recognition \cite{graves13speech}. Despite their superior performance, RNNs are usually considered as ``black-boxes'' which lack transparency, limiting their application in many critical decision-making scenarios \cite{caruana2015intelligible}. The demand for more transparent and intelligible machine learning systems is becoming even more urgent as recent regulations in the European Union require ``the right to explanation'' for algorithms used in individual level predictions \cite{regulation2018general}. \looseness=-1


To address this challenge, a variety of methods have been developed to unveil the inner-workings of deep sequence models through visualizing the changes in hidden states \cite{karpathy2015visualizing,strobelt2018lstmvis}, extracting feature importance \cite{alikaniotis2016automatic, murdoch2017automatic, murdoch2018beyond} and constructing rules that mimic the behavior of RNNs \cite{che2016interpretable}. However, post-hoc explanations can be incomplete or inaccurate in capturing the reasoning process of the original model. Therefore it is often desirable to have models with inherent interpretability in many application scenarios \cite{rudin2018please}. \looseness=-1

We leverage the concept of prototype learning to construct deep sequence model with built-in interpretability. Prototype learning is a form of case-based reasoning \cite{kolodner1992introduction, schmidt2001cased}, which draws conclusions for new inputs by comparing them with a few exemplar cases (i.e. prototypes) in the problem domain \cite{li2018prototype, chen2018prototype-patch}. It is a natural practice in our day-to-day problem-solving process. For example, physicians perform diagnosis and make prescriptions based on their experience with past patients and mechanics predict potential malfunctions by recalling vehicles exhibiting similar symptoms. Prototype learning imitates such human problem-solving process for better interpretability. Recently the concept has been incorporated in convolutional neural networks to build interpretable image classifiers \cite{li2018prototype, chen2018prototype-patch}. However, so far prototype learning is not yet explored for modeling sequential data. \looseness=-1

We propose prototype sequence network (\systemname), which combines prototype learning with variants of RNN to achieve both interpretability and high accuracy for sequence modeling. The RNN as the backbone captures the latent structure of the temporal development. Prediction on a new input sequence is performed based on its similarity to the prototypes in the latent space. For better interpretability, we consider the following criteria in constructing prototypes for explanation: \looseness=-1

\textbf{\textit{Simplicity.}} It is possible to directly use the original sequences in the  data as prototypes, but these sequences may contain irrelevant noises. In our approach, the prototypes can be subsequences of the original training data and contain only the key events determining the output. Shorter prototypes are preferred for presenting the explanation in a more succinct form. \looseness=-1

\textbf{\textit{Diversity.}} Redundant prototypes should be avoided since they add to the complexity of the explanation but do not bring extra performance. Therefore we encourage using a set of prototypes that are sufficiently distinct from each other. The prototypes also give a high-level overview of the original data which can be several magnitudes larger. \looseness=-1

\textbf{\textit{Sparsity.}} For each input it is desirable that only a few prototypes are ``activated'' such that people are not overwhelmed with long and redundant explanations. \looseness=-1

We introduce a novel learning objective which takes the above criteria into consideration and propose a training procedure which iteratively performs gradient descent and prototype projection. For steerable learning, we consider a constrained training process with a number of user-specified prototypes which reflect the experts' intuition and experience in the domain.  \looseness=-1

\systemname~is evaluated on several real-world datasets and it is able to achieve comparable performance with state-of-the-art deep learning techniques. The experiments cover a diverse range of applications including predictive maintenance of automotives, classification of protein sequences, annotation of electrocardiography (ECG) signals and sentiment analysis on customer reviews, demonstrating the general applicability of the method. In each experiment we not only report classification accuracy on training and test data, but also demonstrate intuitive interpretations of the result through concrete case studies and visualizations. We further study the effect of the number of prototypes $k$ and provide guidelines for selecting $k$. Besides that, we also perform studies to explore the effect of including the diversity and the simplicity criteria in the model. \looseness=-1

To further evaluate the interpretability of the prototypes, we conduct a user study on Amazon Mechanical Turk (MTurk) for a sentiment analysis task on customer reviews, the result shows that \systemname~is able to select high quality prototypes that are well-aligned with human knowledge on natural languages for sentiment classification. Finally, we demonstrate that through learning under constraints with user-specified prototypes, the model can be steered to obtain comparable performance with better interpretability. \looseness=-1

The main contribution of this paper is summarized as follows:
\begin{itemize}
    \item A sequence model that learns interpretable representations via sequence prototypes for predictive tasks.
    \item An interaction scheme which allows human experts to incorporate their domain knowledge by validating and updating the learned sequence prototypes.
    \item Experiments on real-world datasets show that \systemname~ achieves comparable performance with the state-of-the-arts while providing analogy based interpretability.
\end{itemize}

The rest of the paper is organized as follows: Section 2 summarizes related work; Section 3 introduces the architecture of \systemname, the learning objective and the training process; Section 4 presents experimental results; Section 5 concludes the paper.

\section{Related Work}

Recently, variants of RNNs including Long-Short Term Memory networks (LSTMs) \cite{lipton2015critical} have been proven to be very effective in modeling sequence data. They have been successfully applied to sentiment analysis \cite{tang15sentiment}, ECG signal classification \cite{kachuee2018ecg}, mortality and disease risk prediction using EHR data \cite{choi2016doctorai, choi2016rnn, harutyunyan2017mt-lstm}, and etc.. \looseness=-1

Despite their impressive performance, these deep learning models consist of complex nonlinear transformations and are often used as ``black boxes'', which leads to trust and fairness issues in many applications \cite{caruana2015intelligible, regulation2018general}. Therefore recently we can observe fast expanding literature in interpretable machine learning. In particular, we review two major approaches to interpretable sequence modeling: 1) post-hoc methods, which derive explanations by looking into existing models 2) sequence models with built-in interpretability.\looseness=-1


\textbf{Post-hoc explanation.} Post-hoc methods unveil the underlying mechanisms of a pre-existing ``black-box'' model. Karpathy \etal~ \cite{karpathy2015visualizing} and Strobelt \etal~\cite{strobelt2018lstmvis} visualize the changes of the internal states in RNNs to understand the roles of the hidden units in retaining temporal information. Another popular approach extracts the importance of each input token in determining the final result \cite{alikaniotis2016automatic, murdoch2017automatic, murdoch2018beyond}. Recently, model distillation is also used to explain deep sequence models \cite{ribeiro2016lime, che2016interpretable, murdoch2017automatic, ribeiro2018anchors}. The basic idea is to build surrogate models that mimic the behavior of the original one. The surrogate models are usually easier to interpret, shedding light into the inner-workings of more complex models. 


\textbf{Sequence models with inherent interpretability.} Built-in interpretability is sometimes desirable since post-hoc explanations usually do not fit the original model precisely\cite{rudin2018please}. Traditional techniques such as decision trees and logistic regression are considered inherently interpretable. However, they lack the capability of modeling complex temporal dependencies in sequential data, thus the performance is often sub-optimal. State-of-the-art researches focus on building models with both interpretability and high accuracy. One example of such models is RETAIN \cite{choi2016retain}, which uses LSTM with attention mechanism for predictive analysis on patient data. The built-in attention highlights the clinic visits and the diagnostic codes that are most critical for the predictions. \looseness=-1

\systemname~  mimics our day-to-day problem-solving process by matching inputs with historical data and producing solutions accordingly. Different from nearest neighbor classifiers used in typical case-based reasoning systems \cite{schmidt2001cased, sun2012supervised}, in our approach only a few selected prototypes are memorized, simplified and used for reasoning. There are several benefits in bringing such sparsity: 1) for different inputs it is easier to compare the predictions as well as their interpretations 2) the learned prototypes give a concise overview of the original data, which can be several magnitudes larger 3) it becomes possible to involve human-in-the-loop to update the prototype interactively such that they can incorporate their domain knowledge to further improve the interpretability of the model. Combining prototype-based reasoning with DNNs is first explored for image classification by Li \etal~and Chen \etal \cite{li2018prototype, chen2018prototype-patch}. In this paper, we incorporate the concept for predictive analysis on sequential data for the first time. \looseness=-1

\textbf{Evaluating interpretability.} There is no universally applicable method to evaluate the interpretability of machine learning models and it is usually use case and model dependent \cite{doshi2017towards}. Quantitative approaches measure the sparsity of the features or the complexity of the model (e.g. number of rules in decision trees). However how these metrics are correlated with human interpretability is still unknown. In this work we evaluate how good the prototypes explain the prediction results based on user studies conducted on MTurk. \looseness=-1

\section{Methodology}

We introduce the architecture of \systemname, formulate the learning objective and describe the training procedure in this section. \looseness=-1

\subsection{\systemname~Architecture}

\begin{figure}[ht]
\includegraphics[width=\linewidth]{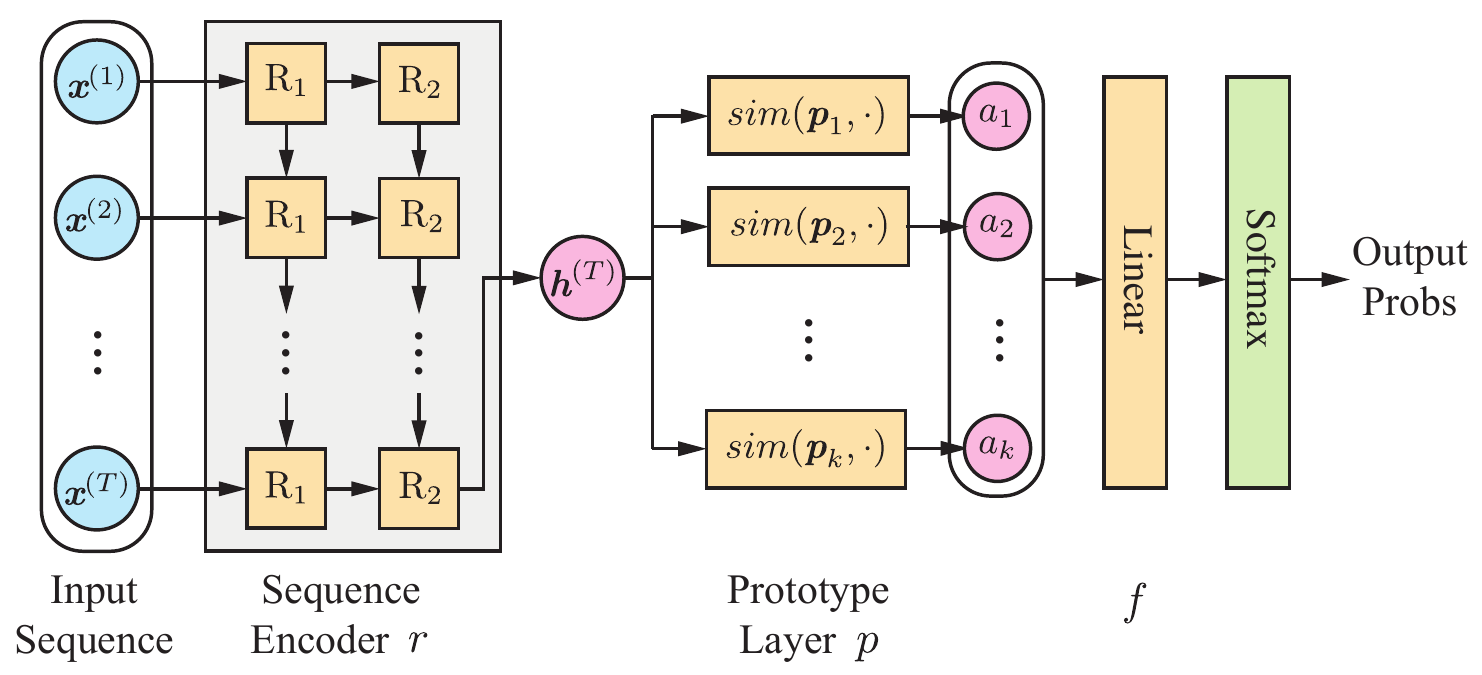}
\vspace{-0.25in}
\caption{The architecture of our proposed \systemname~model. The model consists of three parts, the recurrent sequence encoder network $r$, the prototype layer $p$ that contains $k$ prototypes, the fully connected layer $f$, and a softmax layer for output probabilities in multi-class classification tasks.}
\vspace{-0.15in}
\label{fig:architecture}
\end{figure}

Let $\mathcal{D} = \{((\mathbi{x}^{(t)})_{t=1}^T, y)\}$ be a labeled sequence dataset, where $T$ is the sequence length, $\mathbi{x}^{(t)} \in \mathbb{R}^n$ is the input vector at step $t$, and $y \in \{1, \dots, C\}$ is the label of the sequence. We aim to learn representative prototype sequences (not necessarily exist in the training data) that can be used as classification references and analogical explanations. For a new input sequence, its similarities with each representative sequences are measured in the learned latent space. Then, the prediction of the new instance can be derived and explained by its similar prototype sequences. \looseness=-1

The basic architecture of \systemname~ is similar to the one proposed by Li \etal \cite{li2018prototype}. As shown in \autoref{fig:architecture}, the model consists of three components: a sequence encoder $r$, a prototype layer $p$, and a fully connected layer $f$. \looseness=-1

For a given input sequence $(\mathbi{x}^{(t)})_{t=1}^T$, the sequence encoder $r$ maps the entire sequence into a single embedding vector with fixed length $\mathbi{e} = r((\mathbi{x}^{(t)})_{t=1}^T)$, $\mathbi{e} \in \mathbb{R}^m$. 
The encoder could be any backbone sequence learning models \eg, LSTM, Bidirectional LSTM (Bi-LSTM) or GRU. 
In our experiments, the hidden state at the last step, $\mathbi{h}^{(T)}$, is used as the embedding vector. \looseness=-1

The prototype layer $p$ contains $k$ prototype vectors $\mathbi{p}_i \in \mathbb{R}^m$, which have the same length as $\mathbi{e}$. The layer scores the similarity between $\mathbi{e}$ and each prototype $\mathbi{p}_i$. 
In previous work \cite{li2018prototype}, the squared $L_2$ distance, $d_i^2 = \|\mathbi{e} - \mathbi{p}_i\|_2^2$, is directly used as the output of the layer.  
To improve interpretability, we compute the similarity using:
$$a_i = \exp(-d_i^2),$$
which converts the distance to a score between 0 and 1. Zero can be interpreted as the sequence embedding $\mathbi{e}$ being completely different from the prototype vector $\mathbi{p}_i$, and one means they are identical. \looseness=-1

With the computed similarity vector $\mathbi{a} = p(\mathbi{e})$, the fully connected layer computes $\mathbi{z} = \mathbi{W}\mathbi{a}$, where $\mathbi{W}$ is a $C \times k$ weight matrix and $C$ is the output size (\ie, the number of classes in classification tasks). To enhance interpretability, we constrain $\mathbi{W}$ to be non-negative. For multi-class classification tasks, a softmax layer is used to compute the predicted probability: $\hat{\mathbi{y}_i} = \exp(z_i)/\sum_{j=1}^C \exp(z_j)$.  \looseness=-1



\subsection{Learning Objective}

Our goal is to learn a \systemname~that is both accurate and interpretable. For accuracy, we minimize the cross-entropy loss on training set:
$CE(\Theta, \mathcal{D}) = \sum_{((\mathbi{x}^{(t)})_{t=1}^T, \mathbi{y}) \in \mathcal{D}}
{\mathbi{y} \log (\hat{\mathbi{y}}) + (1 - \mathbi{y})\log (1-\hat{\mathbi{y}})},$
where $\Theta$ is the set of all trainable parameters of the model. \looseness=-1

\vskip 0.05in
\noindent\textbf{Diversity.} In our experiments, we found that when the number of prototypes $k$ is large (\ie, over two or three times the number of classes), the training would often result in a number of similar or even duplicate prototypes (\ie, some prototypes are very close to each other in the latent space). It would be confusing to have multiple similar prototypes in the explanations and also inefficient in utilizing model parameters. We prevent such phenomenon through a diversity regularization term that penalizes on prototypes that are close to each other:
$$R_{d}(\Theta) = \sum_{i=1}^k \sum_{j=i+1}^k \max\left(0, d_{min} - \|\mathbi{p}_i - \mathbi{p}_j\|_2 \right)^2,$$
where $d_{min}$ is a threshold that classifies whether two prototypes are close or not. We set $d_{min}$ to 1.0 or 2.0 in our experiments. $R_d$ is a soft regularization that exerts a larger penalty on smaller pairwise distances. By keeping prototypes distributed in the latent space, it also helps produce a sparser similarity vector $\mathbi{a}$. \looseness=-1

\vskip 0.05in
\noindent\textbf{Sparsity and non-negativity.} In addition, to further enhance interpretability, we add $L_1$ penalty on the fully connected layer $f$, and constrain the weight matrix $\mathbi{W}$ to be non-negative. The $L_1$ sparsity penalty and non-negative constraints on $f$ help to learn sequence prototypes that have more unitary and additive semantics for classification. \looseness=-1

\vskip 0.05in
\noindent\textbf{Clustering and evidence regularization.} To improve interpretability, Li \etal \cite{li2018prototype} also proposed two regularization terms to be jointly minimized, the clustering regularization $R_{c}$ and the evidence regularization $R_{e}$. $R_c$ encourages a clustering structure in the latent space by minimizing the squared distance between an encoded instance and its closest prototype:
$$R_{c}(\Theta, \mathcal{D}) = \sum_{(\mathbi{x}^{(t)})_{t=1}^T \in \mathcal{X}} \min_{i=1}^k \left\|r\left((\mathbi{x}^{(t)})_{t=1}^T\right) - \mathbi{p}_i\right\|_2^2,$$
where $\mathcal{X}$ is the set of all sequences in the training set $\mathcal{D}$. The evidence regularization $R_e$ encourages each prototype vector to be as close to an encoded instance as possible:
$$R_{e}(\Theta, \mathcal{D}) = \sum_{i=1}^k \min_{(\mathbi{x}^{(t)})_{t=1}^T \in \mathcal{X}} \left\| \mathbi{p}_i - r\left((\mathbi{x}^{(t)})_{t=1}^T\right) \right\|_2^2.$$

\vskip 0.05in
\noindent\textbf{Full objective.} To summarize, the loss that we are minimizing is:
\begin{equation}
\begin{aligned} 
Loss(\Theta, \mathcal{D}) =&\ CE(\Theta, \mathcal{D}) + \lambda_c R_{c}(\Theta, \mathcal{D}) + \lambda_e R_{e}(\Theta, \mathcal{D}) 
\\
&+ \lambda_d R_{d}(\Theta, \mathcal{D}) + \lambda_{l_1} \| \mathbi{W}\|_1,
\end{aligned}
\end{equation}
where $\lambda_c$, $\lambda_e$, $\lambda_d$ and $\lambda_{l_1}$ are hyperparameters that control the strength of the regularizations. The configuration of these hyperparameters largely depends on the nature of the data and can be selected through cross-validation. For each experiment in Section 4, we provide the hyperparameter settings. \looseness=-1


\subsection{Optimizing the Objective}

We use stochastic gradient descent (SGD) with mini-batch to minimize the loss function on training data. 
In this section, we mainly discuss the prototype projection and simplification techniques that we used to learn simple and interpretable prototypes. 
\looseness=-1


\vskip 0.05in
\noindent\textbf{Prototype projection.} Since the prototype vectors $\mathbi{p}_i$ are representations in the latent space, they are not readily interpretable. Li \etal\cite{li2018prototype} proposed to jointly train a decoder that translates the latent space to the original input space to make prototypes interpretable. However, the decoder may not necessarily decode prototypes to meaningful sequences. Instead of using a decoder, we design a projection step during training that assigns $\mathbi{p}_i$ with their closest sequence embedding in the training set:
\begin{equation}
\mathbi{p}_i \leftarrow \argmin_{\mathbi{e} \in r(\mathcal{X})} \left\|\mathbi{e} - \mathbi{p}_i \right\|_2.
\label{eq:projection}
\end{equation}
Each prototype vector $\mathbi{p}_i$ is then associated with a \textit{prototype sequence} in the input space. The projection step is only performed every few training epochs (we set to 4 in our experiments) to reduce computational cost. Compared with the original prototype network \cite{li2018prototype}, the projection step saves the efforts of jointly training a sequence auto-encoder, which is computationally expensive. It also assures each prototype to be an observed sequence, which guarantees that the prototypes are meaningful in the real world. \looseness=-1

\vskip 0.05in
\noindent\textbf{Interpretation with prototypes.} \systemname~is readily explainable by consulting the most similar prototypes. When making predictions based on a new input sequence, the explanation can be generated along with the inference procedure. A prediction could be explained by a weighted addition of the contribution of the most similar prototypes: \looseness=-1

\noindent\begin{tabular}{ll}
Input: & pizza is good but service is extremely slow \\
Prediction: & Negative \\
Explanation: & \ \ \ 0.69 * good food but worst service (Negative 2.1) \\
& + 0.30 * service is really slow (Negative 1.1) 
\end{tabular}
The factors in front of the prototype sequences are the similarities between the input and the prototypes. At the end of each prototype shows its associated weights $\mathbi{w}_i$. The weights can be interpreted as the model's confidence on the possible labels of the prototype. \looseness=-1


\vskip 0.05in
\noindent\textbf{Prototype simplification.} Although the prototypes are already readable after projecting to observed sequences in the training data, it may still be difficult to comprehend a prototype sequence if it contains insignificant or irrelevant noisy events. \looseness=-1

Next, we introduce a procedure to simplify the projected prototype sequences. That is, instead of projecting a prototype to a complete observed sequence, we project it to a subsequence containing the critical events. The projection step (\autoref{eq:projection}) now becomes: \looseness=-1
\begin{equation}
\begin{gathered}
    \mathbi{p}_i \leftarrow r(seq_i), \\
    seq_i = \argmin_{seq \in sub(\mathcal{X})} 
    \left(
        \left\|r(seq) - \mathbi{p}_i \right\|_2
    \right),
    \label{eq:projection-simplification}
\end{gathered}
\end{equation}
where $sub(\mathcal{X})$ is the set of all possible subsequences of the data in $\mathcal{X}$, $|\cdot|$ computes the effective length of the subsequence. Note that the complexity of the above operation is $O(2^TN)$, where $N$ is the size of training set and $T$ is the maximum length of the sequences in $\mathcal{X}$. The cost of the brute-force computation grows exponentially with $T$, which is unacceptable even for relatively short sequences. \looseness=-1

We use beam search to find an approximate solution \cite{rich1991artificial}. Beam search is a greedy breadth-first search algorithm which only keeps $w$ best candidates in each iteration. $w$ is called the beam width. The algorithm first selects $w$ closest candidate sequences to prototype $\mathbi{p}_i$. Then it generates all the possible subsequences which can be obtained by removing one event from any of the $w$ candidates. The score in \autoref{eq:projection-simplification} is calculated for each subsequence. The $w$ subsequences with the minimum scores are then kept as candidates to continue the search in the next iteration. The subsequence with the minimum score is the output. The complexity of the algorithm is now $O(w\cdot T^2N)$. We use $w=3$ in our experiments. \looseness=-1

\subsection{Refining \systemname~with User Knowledge}\label{sec:method-refine}

\begin{figure}[ht]
\includegraphics[width=\linewidth]{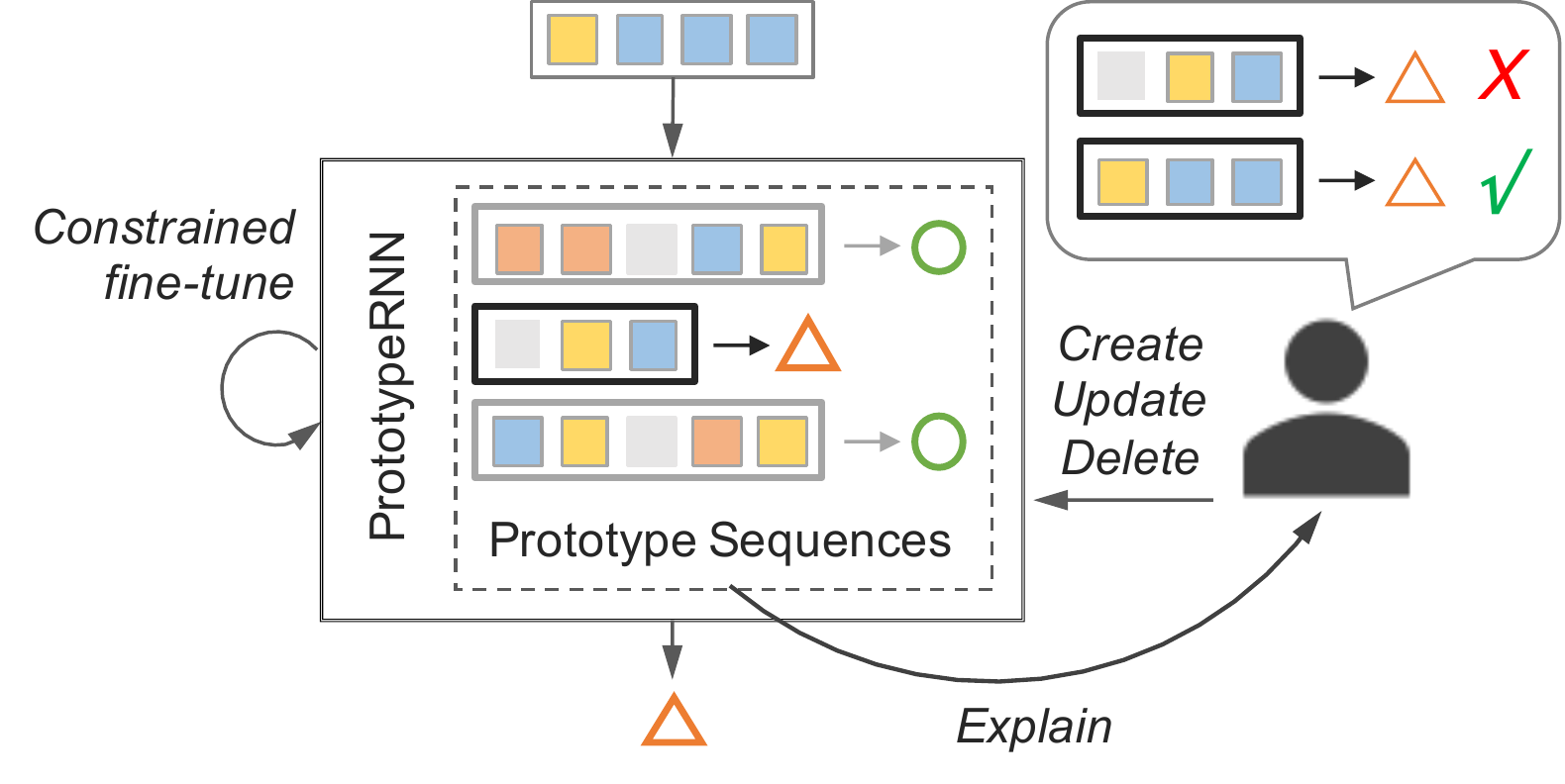}
\vspace{-0.25in}
\caption{A user verifying and refining a \systemname~with his/her knowledge and observation on the model output.}
\vspace{-0.15in}
\label{fig:interaction-scheme}
\end{figure}

Next, as illustrated in \autoref{fig:interaction-scheme}, we discuss how users can refine a \systemname~for better interpretability and performance by validating and updating the prototypes, esp. when they have certain expertise or knowledge in the problem domain. Allowing users to validate and interact with the prototypes can also increase their understanding of the model and the data, which is the foundation of user trust \cite{ribeiro2016lime}. \looseness=-1

We assume that the knowledge of a user can be explicitly expressed in the form of input-output patterns which the user recognizes as significant or typical in the domain (\eg, ``food is good'' is typically a review with ``positive'' sentiment). These patterns can be regarded as the ``prototypes'' that the user learned from his/her past experiences. The refinement can thus be done by incorporating user-specified prototypes as constraints in the model. \looseness=-1

Based on the users' past knowledge and observation on the model outputs, there are three types of possible operations that they can apply to the model: \textit{create} new prototypes, \textit{revise} or \textit{delete} existing prototypes. After changes are committed, the model is fine-tuned on the training data to reflect the change. \looseness=-1

When fine-tuning the model the prototypes should be fixed to reflect the users' constraints. Therefore we make the following revisions to the optimization process described in Section 3.3: 1) instead of updating the latent prototype vectors $\mathbi{p}_i$ in the gradient descent step, we use the updated sequence encoder $r$ in each iteration to directly set $\mathbi{p}_i = r(seq_i)$; 2) the prototype projection step is skipped. After fine-tuning, the sequence encoder $r$ learns better representations of the data. The user can verify the updated results and repeat the process until he/she is satisfied with the result.\looseness=-1

%
%

\section{Experimental Evaluation}

In this section, we evaluate \systemname~for classification tasks on four real-world sequence datasets. Besides performance metrics, we also evaluate the interpretability of \systemname~with both qualitative case studies and quantitative experiments with human users. We also perform ablation studies to understand how the following factors affect the performance: the prototype number $k$, the diversity regularization term and the prototype simplification step. \looseness=-1

We implemented \systemname\footnote{Code available for research purposes at \texttt{\systemurl}} using PyTorch\footnote{https://pytorch.org/}. We use stochastic gradient descent (SGD) for the training of all models. We clip the $L_2$ norm of the gradients at 5 to prevent exploding gradient during the training \cite{pascanu13rnn_training}. The learning rate is set to 1.0 for the first 10 epochs, and is decayed with a factor of 0.85 for each epoch afterwards. \looseness=-1


\subsection{Case Study 1: Predictive Diagnostics based on Vehicle Fault Log Data}
Today's vehicles have complex interconnected modules and the faults usually have a significant history of development over a vehicle's lifetime. Fault logs collected from cars can therefore be used to understand the typical development paths of the problems and support predictive diagnostics. The fault log of each vehicle can be modeled as a sequence of events. Each event corresponds to one or multiple faults that happen at the same time. Each fault is described with a five-digit Diagnostic Trouble Code (DTC) which is standard across different makes and models. With \systemname, we aim to predict the risk of faults (\ie~ DTCs) for a vehicle in the future using its historical DTC logs. We encode an event as a multi-hot vector since multiple faults could occur at the same time. The input at each step is therefore a binary vector $\mathbi{x}^{(t)} \in \{0, 1\}^n$  and each element in the vector indicates if a particular fault has occurred. The problem is formulated as multi-label classification to predict the risk of different DTCs. The softmax layer is replaced with a sigmoid layer to compute the output probabilities. \looseness=-1

In total there are 12k vehicle fault sequences containing 393 different types of DTCs. We train the classifier to predict the top 92 DTCs which have occurred more than 100 times in the dataset. The sequences have an average length of 2.31. The dataset is split into 7.2k training, 2.4k validation, and 2.4k test set. We train a \systemname~ with an LSTM encoder (1 layer, 50 hidden units) and 100 prototypes. We set $\lambda_{l_1} = 1.0, \lambda_e=0.1, \lambda_c=0.01, \lambda_d=0.01, d_{min} = 1.0$ during the training. For prototype simplification, we set the beam width $w=3$. We use recall at 5 (Recall@5) and mean average precision at 5 (MAP@5) as performance measures. \looseness=-1

We compare the performance of \systemname~ with a standard LSTM with the same number of layers and hidden units (\autoref{tab:dtc-performance}). Both models are trained for 24 epochs with a batch size of 64. 
\looseness=-1

\begin{table}[bth]
  \vspace{-0.05in}
  \caption{Performance on vehicle fault risk prediction.}
  \vspace{-0.05in}
  \label{tab:dtc-performance}
	\centering%
    \begin{tabular} {c c c}
    \hline
    Model & Recall@5 (\%) & MAP@5 \\
    \hline
    \systemname & 0.473 & 0.759 \\
    LSTM & 0.479 & 0.751 \\
    \hline
    \end{tabular}
    \vspace{-0.05in}
\end{table}

\begin{figure}[hbt]
\includegraphics[width=0.9\linewidth]{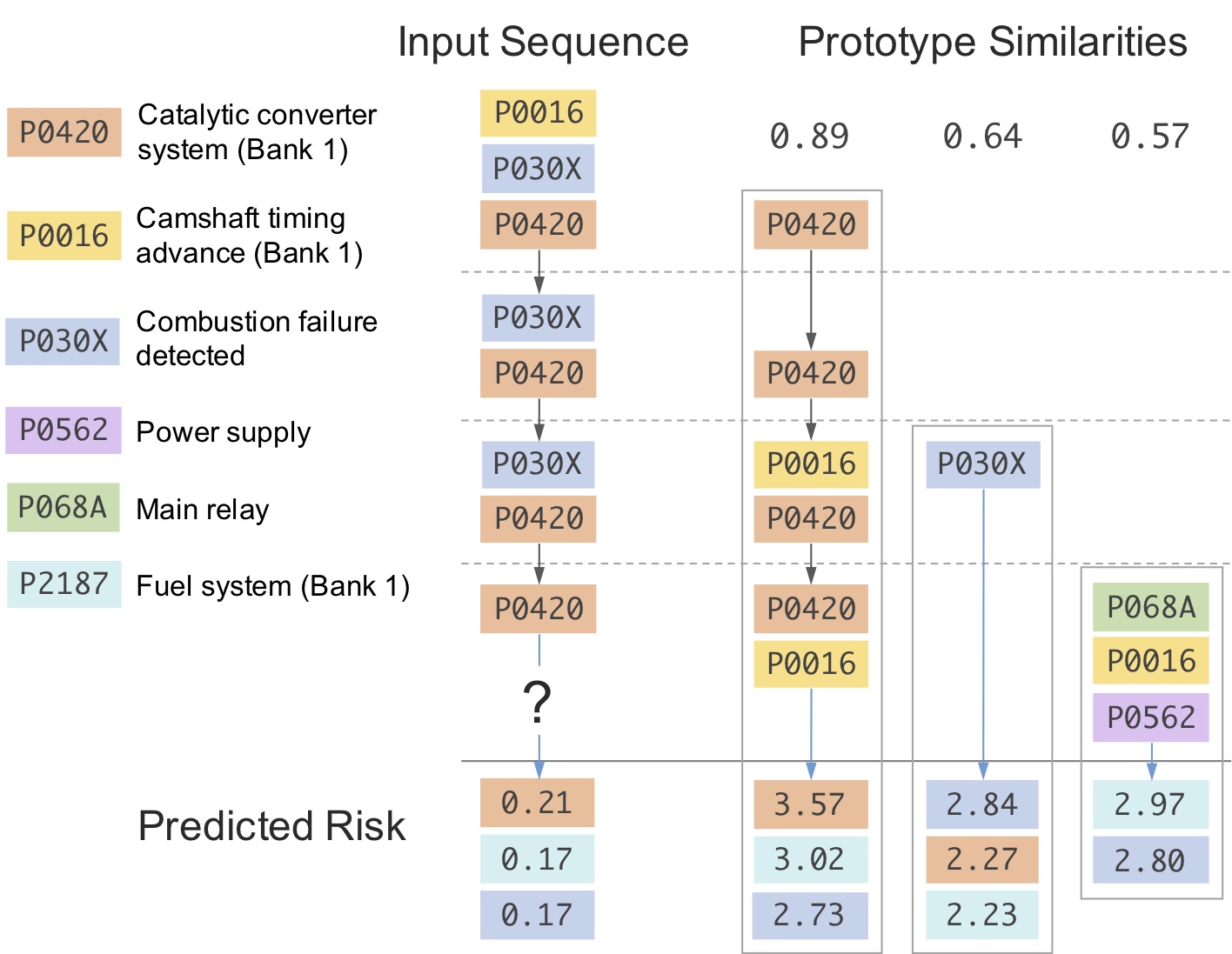}
\vspace{-0.10in}
\caption{An input sequence and the similarity scores with its closest prototypes. The weights $\mathbi{w}_i$ are visualized at the bottom as the outcome of the prototype sequences.}
\vspace{-0.05in}
\label{fig:vehicle-fault-prototype}
\end{figure}

An example prediction of the model on an input fault log sequence is shown in \autoref{fig:vehicle-fault-prototype}. The input sequence shows a recurring sequence consisting of ``P030X'' and ``P0420'', while the model predicts a relatively high risk (0.17) of  ``P2187'' which has not occurred before. ``P2187'' indicates a problem of the fuel system in the engine at Bank 1. With the given explanation, we can see that there are three prototypes that match different aspects of the input sequence. All of the three prototypes indicate a high risk of ``P2187'', which explains the reasons of the prediction. A mechanic could utilize the model to predict potential future problems and ground the predictions on exemplar cases. The entire set of prototypes in the model provides an overview of all the fault development paths, which can help manufacturers identify systematic issues and develop preventive maintenance strategies. \looseness=-1

\subsection{Case Study 2: Sentiment Analysis}\label{sec:exp-yelp}

We also evaluate \systemname~using a sentiment classification task on text data. We use the reviews of restaurants in the Yelp Open Dataset\footnote{http://www.yelp.com/dataset}. 
Each review is tokenized into a sequence of words using NLTK\footnote{https://www.nltk.org/}. We only use reviews that are less than 25 words in the experiments (106k reviews in total) since in later user study (Section 4.6) shorter sentences are easier for humans to read and compare. We use the stars (one to five) given with the reviews as labels and conduct experiments on both \textit{fine-grained} (5-class) and \textit{binary} (positive=rating $\geq$ 3) classifications. The dataset is split into 60\% training, 20\% validation, and 20\% test set.  \looseness=-1

\begin{table}[hbt]
    \vspace{-0.05in}
  \caption{Average accuracy (\%) of LSTM, \systemname~ and ResNet on 10 random train-test splits of Yelp Reviews. Numbers in parentheses are standard deviations.}
  \vspace{-0.1in}
  \label{tab:yelp-performance}
	\centering%
    \begin{tabular} {c c c}
    \hline
    Model & Binary & Fine-grained \\
    \hline
    \systemname & 95.5 (0.1) & 60.7 (0.2) \\
    \systemname\textsubscript{Bi-LSTM} & 95.5 (0.2) & 61.0 (0.3) \\
    \hline
    LSTM & 95.6 (0.1) & 60.7 (0.3) \\
    Bi-LSTM & 95.7 (0.1)  & 61.1 (0.3) \\
    \hline
    ResNet & 95.3 (0.3) & 60.6 (0.4) \\
    \hline
    \end{tabular}
    \vspace{-0.05in}
\end{table}

As shown in \autoref{tab:yelp-performance}, we report the accuracy of \systemname, \systemname~with Bi-LSTM encoder, LSTM, Bi-LSTM, and ResNet on both validation and test set. All LSTMs have 2 layers, with 100 hidden units per layer. The ResNet contains 7 residual blocks similar to the architecture mentioned in \cite{he2016deep}. We apply a dropout rate of 0.8 during training. The initial number of prototypes is set to 100 and 200 in binary and fine-grained classification tasks respectively. The result shows that our model can learn interpretable representations while achieving similar, though slightly lower performance compared with the state-of-the-art bi-directional LSTMs.  \looseness=-1 

\begin{figure}[htb]
\includegraphics[width=\linewidth]{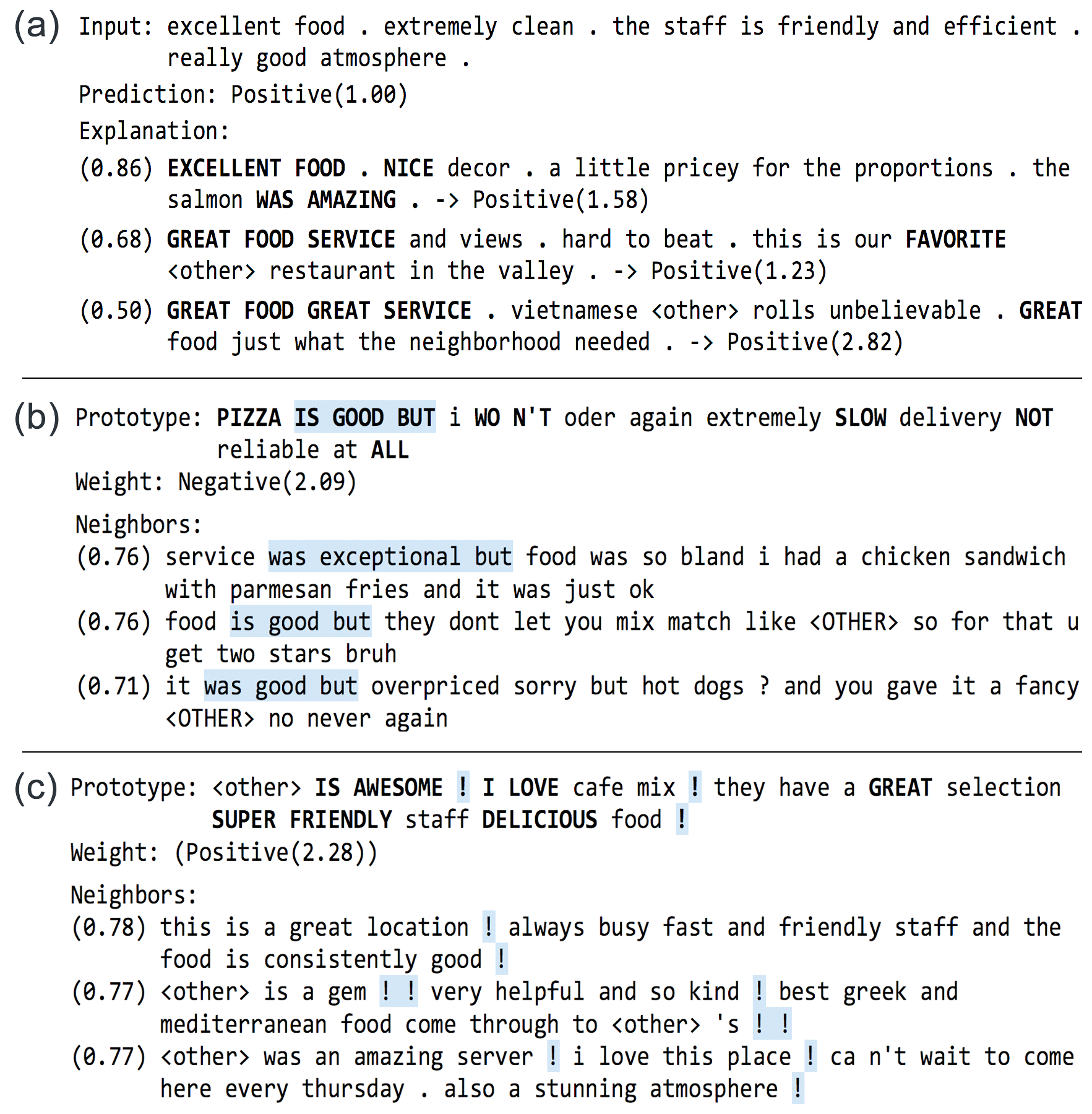}
\vspace{-0.25in}
\caption{Examples of binary sentiment classification on Yelp Reviews. (a) The generated explanation of a  prediction. (b) and (c): Prototypes and their neighboring sentences. Similar patterns between prototypes and neighboring sequences are manually highlighted in blue. The numbers show the similarities between the input and prototypes. Bold and uppercase texts show the simplified prototype sequences. }
\vspace{-0.10in}
\label{fig:yelp-prototype}
\end{figure}

\autoref{fig:yelp-prototype} (a) shows the explanation of an example sentiment analysis result. The referenced prototypes show the different aspects of a good restaurant --- good food and service. Some other prototype sequences and their neighboring sequences are presented in \autoref{fig:yelp-prototype} (b)(c). We can see that some prototypes represent frequent short phrases that are sentimentally significant. Some prototypes capture long-range semantics, such as the transition of sentiments via contrastive conjunctions (\eg, but in \autoref{fig:yelp-prototype} (b)). We also discovered some interesting sequential ``patterns'' of how people express their sentiments. For example, a typical way of expressing positive sentiment is through multiple short compliments ended with exclamation marks (\autoref{fig:yelp-prototype} (c)). Note that the input sequences and the prototype sequences are matched through a learned distance measure through an LSTM rather than strict pattern matching.  \looseness=-1

\subsection{Case Study 3: UniProtKB Protein Sequence Classification}\label{sec:case-protein}

We evaluate \systemname~ in the biology domain using the UniProtKB database. The database contains 558,898 protein sequences manually annotated and reviewed. Protein sequences are composed of 20 standard amino acids and can be grouped into families. Proteins in a family descend from a common ancestor and typically have similar functions and 3D structure. We investigate whether \systemname~ can learn the sequential similarity within families.  \looseness=-1

\begin{figure}[bt]
\includegraphics[width=\linewidth]{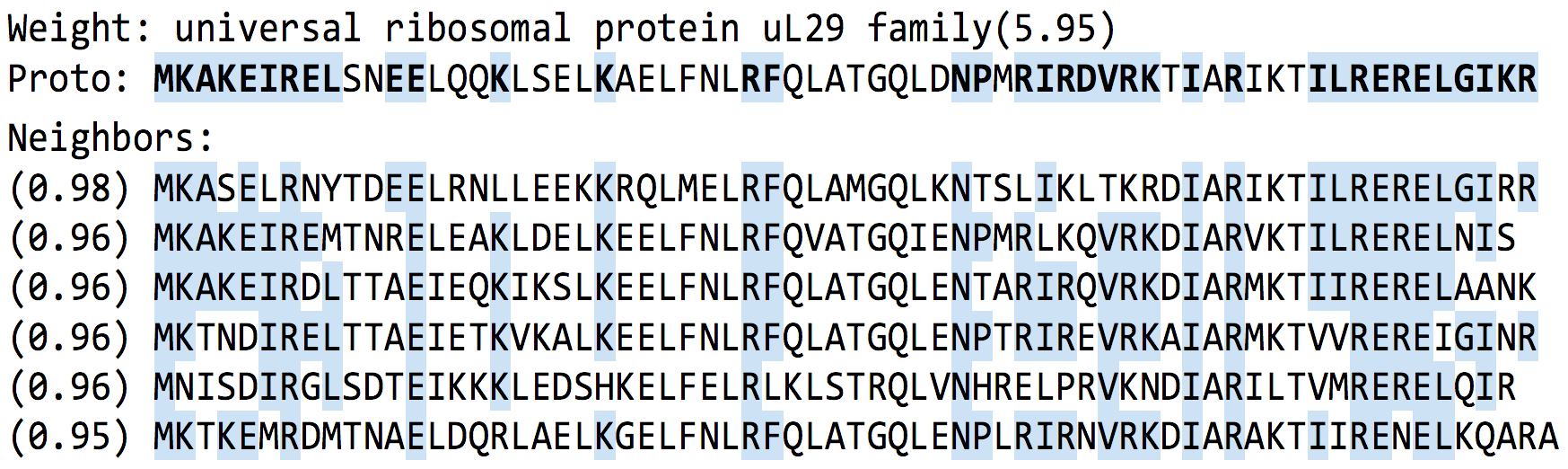}
\vspace{-0.20in}
\caption{A prototype protein sequence and its neighboring sequences in the test data. The bold and highlighted characters show the simplified prototype subsequence. The matching subsequences in the neighbors are highlighted in blue.}
\vspace{-0.10in}
\label{fig:protein-prototype}
\end{figure}

We clip the sequences with a maximum length of 512 and aim to classify the top 100 families ranked by their size. The sequences are split into 62k train and 19k test set. We set $\lambda_{l_1} = 1.0, \lambda_e = 0.1, \lambda_c=0, \lambda_d = 0.01, d_{min} = 1.0$ and train a \systemname~consists of a Bi-LSTM (2 layer $\times$ 50 hidden units) and 200 prototypes. 
We trained the \systemname~ for 40 epochs with a batch size of 64. 
In 10 train-test splits, the model scores an average accuracy of 97.0\% ($SD= 0.2\%$) on the test set. Its accuracy is slightly lower than a Bi-LSTM (97.4\%, $SD= 0.2\%$) and slightly higher than a ResNet with seven residual blocks (96.7\%, $SD= 0.3\%$). However, the \systemname~learns interpretable representations that reveal the significant subsequences for a family of proteins. An example is shown in \autoref{fig:protein-prototype}. \looseness=-1

\subsection{Case Study 4: ECG Signal Classification}

We investigate whether \systemname~ can be extended to learn meaningful prototypes in real-valued time series using the MIT-BIH Arrhythmia ECG dataset\footnote{https://www.physionet.org/physiobank/database/mitdb/}. ECG is widely used in medical practices to monitor cardiac health. Correct categorization of the waveforms is critical for proper diagnosis and treatment. In the dataset, each signal consists of heartbeats annotated by at least two cardiologists. We downsample the ECG signals to 125Hz and split them into annotated heartbeats according to the protocol proposed by Kachuee \etal\cite{kachuee2018ecg}. The annotations are mapped into five groups as suggested by AAMI\cite{AAMI}: Normal (N), Supraventricular Ectopic Beat (SVEB), Ventricular Ectopic Beat (VEB), Fusion Beat (F) and Unknown Beat (Q). The training and test set contain 87k and 21k sequences respectively.  \looseness=-1

Instead of discretizing the time series data into event sequences \cite{bertens2016sumsequence}, we directly use LSTM to encode the real-valued sequence. We set $\lambda_{l_1}=0.1$, $\lambda_e = 1.0$, $\lambda_d=0.01$, $d_{min}=2.0$, dropout rate to 0.1, and train a \systemname~ with a Bi-LSTM encoder (32 hidden units $\times$ 3 layers) and 30 prototypes. The training runs for 36 epochs with a batch size of 128 and prototype simplification is not applied. After removing prototypes with small weight ($\max(\mathbi{w}_i) < 0.1 \max(\mathbi{W})$), we obtain a model with 23 prototypes.  \looseness=-1

\begin{figure}[htb]
\includegraphics[width=0.95\linewidth]{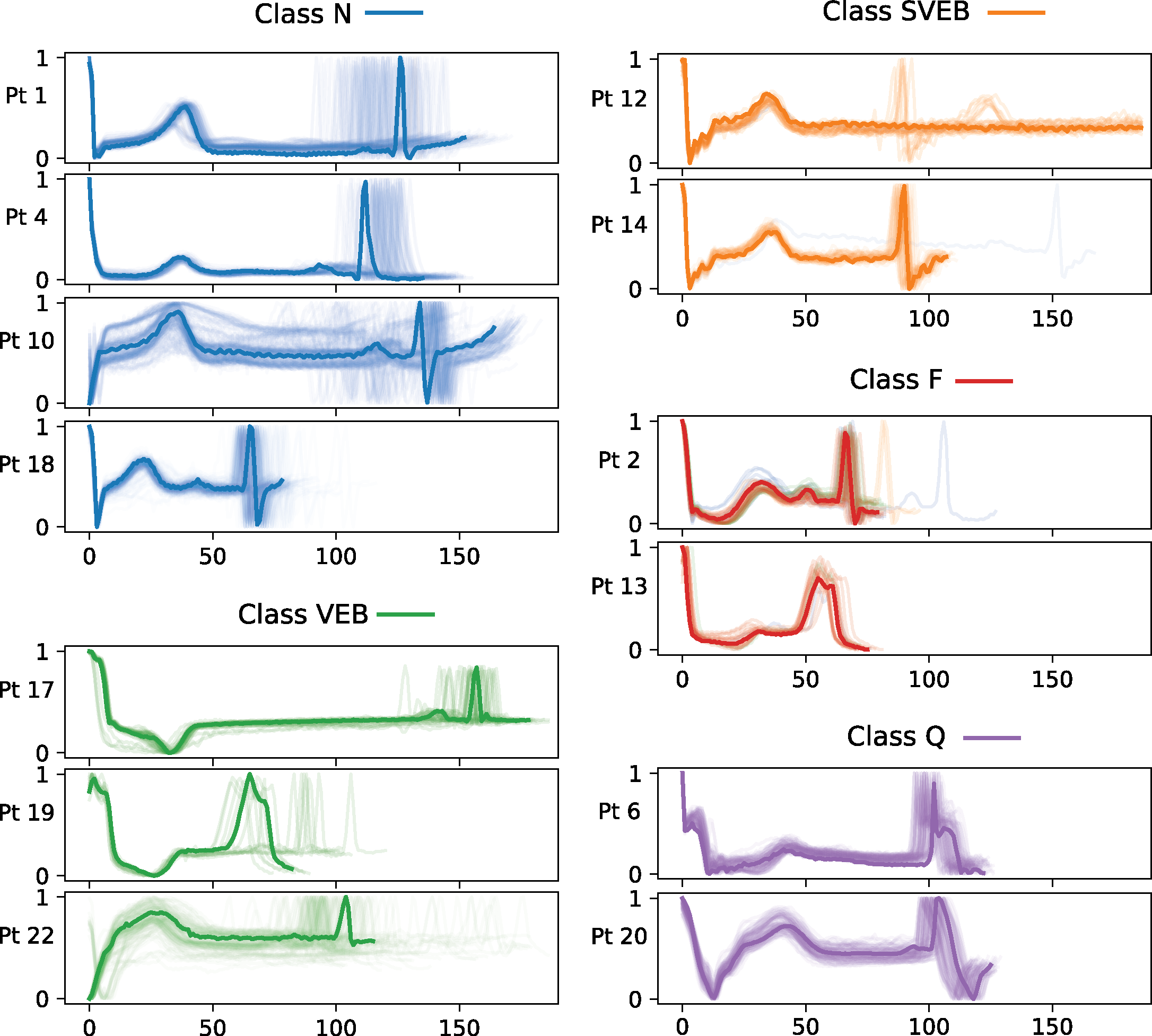}
\vspace{-0.15in}
\caption{Example prototypes of the heartbeat signals. The prototypes are shown as bold lines and colored with their associated label. The color encodes the label of the signals. Transparent lines represent heartbeats in the test set, displayed along with their closest prototypes.}
\vspace{-0.05in}
\label{fig:ecg-prototypes}
\end{figure}

A few selected prototypes are shown in \autoref{fig:ecg-prototypes}. We can see that the \systemname~successfully learned a few prototypes for each class. Prototype 12 shows a characteristic junctional escape beat (belonging to the SVEB group), which shows a long flat line corresponding to the dropped beat. Prototype 17 shows a premature ventricular contraction beat with a strong contraction and a long pause between ventricular contraction.
This demonstrates the capability of \systemname~in learning meaningful representations on ECG time series data, which has also been verified by two independent cardiologists.  \looseness=-1

We also compared our model with the state-of-the-art models for classification of ECG heartbeats. The result is summarized in \autoref{tab:ecg-performance}. We can see that \systemname~has comparable performance to LSTM, and even slightly better accuracy than Residual CNN \cite{kachuee2018ecg}. Our model can present verifiable and understandable prototypes which are very useful in the healthcare domain. In practice, the most similar prototypes can be presented side-by-side with the automatic annotations of the ECG signals for explanation.  \looseness=-1

\begin{table}[hbt]
  \vspace{-0.05in}
  \caption{Performance on MIT-BIH Arrythmia ECG heartbeats classification. }
  \vspace{-0.05in}
  \label{tab:ecg-performance}
	\centering%
    \begin{tabular} {c c c c}
    \hline
    Model & Acc. (\%) & Avg. Precision & Avg. Recall \\
    \hline
    \systemname\textsubscript{Bi-LSTM} & 97.7 & 85.0 & 92.6 \\
    Bi-LSTM & 98.0 & 90.0 & 89.3 \\
    Residual CNN \cite{kachuee2018ecg} & 97.5 & 86.3 & 90.0 \\
    \hline
    \end{tabular}
    \vspace{-0.05in}
\end{table}

\subsection{Ablation Studies}

\subsubsection{Choosing the Number of Prototypes $k$.}
\begin{figure}[hbt]
\includegraphics[width=0.9\linewidth]{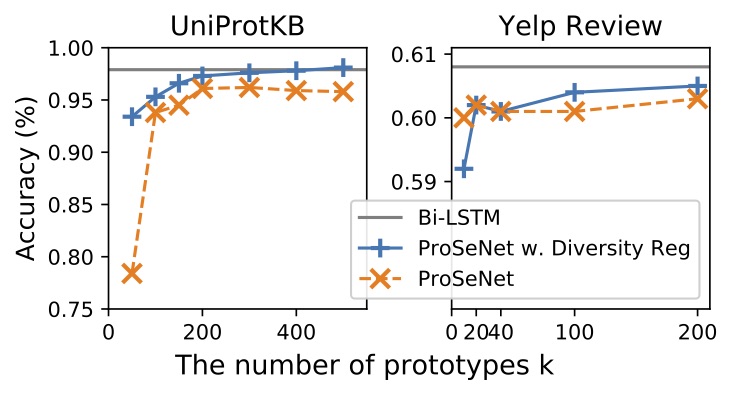}
\vspace{-0.10in}
\caption{The influence of the prototype number and the diversity regularization $R_d$ ($\lambda_d=0.01$) on the performance.}
\vspace{-0.1in}
\label{fig:n_pts-performance}
\end{figure}

We investigate how would the number of prototypes, $k$, influence the performance of \systemname~ using UniProtKB and Yelp Review data. Using the same hyperparameter configuration as in Section~\ref{sec:exp-yelp} and Section~\ref{sec:case-protein}, we train a series of \systemname s with different $k$. 
As shown in the blue lines in \autoref{fig:n_pts-performance}, the accuracy first improves dramatically as $k$ increases. Then the increasing slope quickly flattens after $k$ exceeds 100 for UniProtKB and 40 for Yelp Reviews. 

\begin{figure}[htb]
\includegraphics[width=0.95\linewidth]{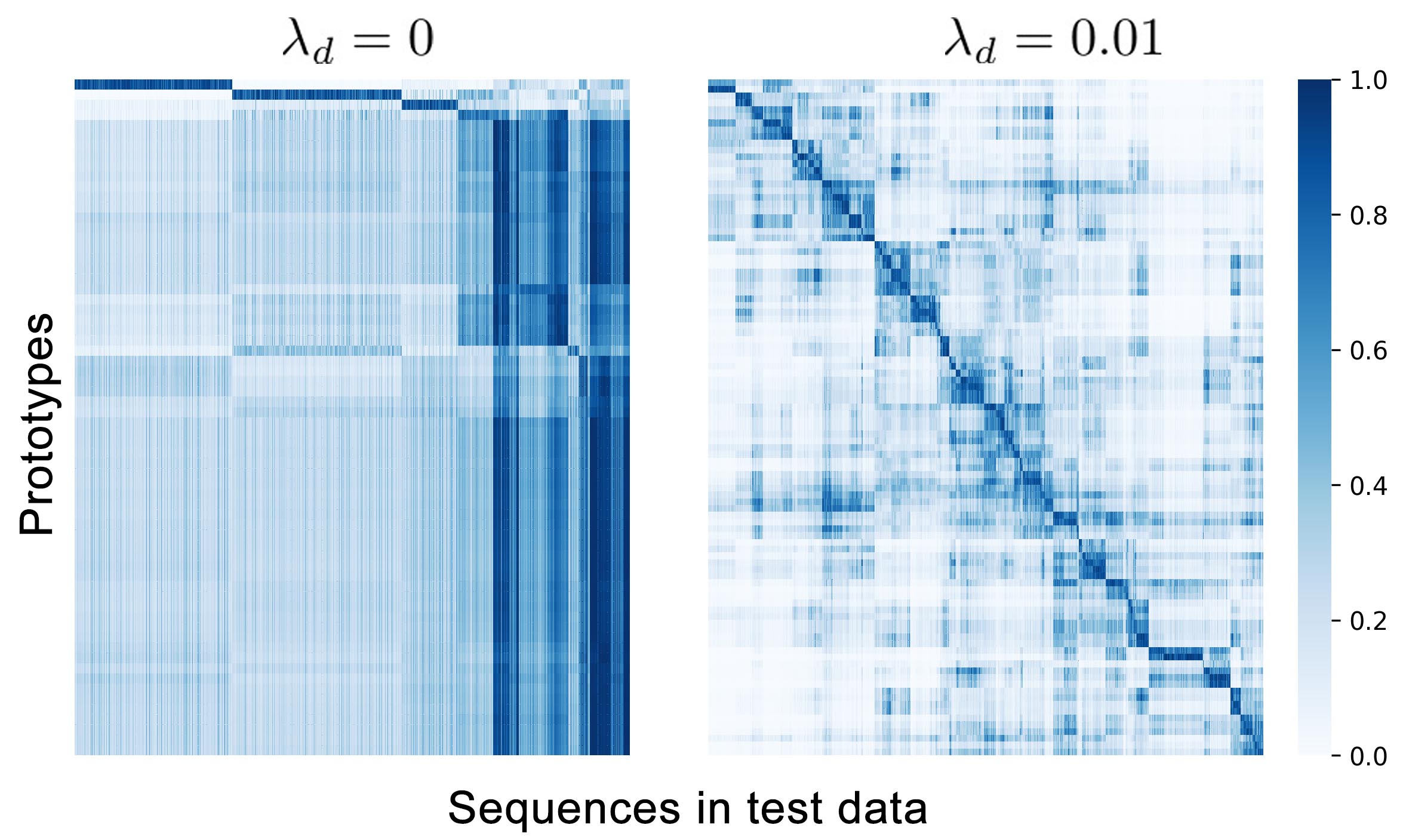}
\vspace{-0.15in}
\caption{The influence of the diversity regularization term $R_d$ on the diversity and sparsity of \systemname. The heatmaps show the similarities between prototypes and test sequences on the Yelp Reviews.}
\vspace{-0.20in}
\label{fig:distinct-heatmap}
\end{figure}

An immediate question is: which $k$ should we use? As $k$ increases, the accuracy improves, however it will become more difficult to comprehend and differentiate such a large number of prototypes. Thus, there is a trade-off between accuracy and interpretability. In practice, since increasing $k$ after a certain threshold only brings marginal improvement to the performance, one possible strategy is to first start from a small $k$ (\eg, $k = C$ to assume one prototype per class) and gradually increase $k$ until the performance improvement falls below a certain threshold.  \looseness=-1

\subsubsection{Effect of the Diversity Regularization Term $R_d$.}
To study the effect of the diversity regularization term, we removed the term by setting $\lambda_d = 0$ and run another set of experiments with varying prototype numbers. The result is also plotted in \autoref{fig:n_pts-performance}. We can observe that the performance on UniProtKB is consistently lower without $R_d$ for different settings of $k$. $R_d$ also positively affects the performance on Yelp Reviews for larger $k$s.  \looseness=-1

We further examine the impact of $R_d$ by plotting the similarity scores between the prototypes and test sequences as heatmaps for two \systemname s with 100 prototypes (\autoref{fig:distinct-heatmap}) on Yelp Review data. Without diversity regularization ($\lambda_d = 0$), most of the rows have similar horizontal patterns in the heatmap (\autoref{fig:distinct-heatmap} left), indicating near-duplicate prototypes. With $\lambda_d=0.01$ the similarity heatmap is much sparser and more diagonal, showing that the prototypes are more diverse and evenly distributed in the latent space. \looseness=-1

\subsubsection{Performance of Prototype Simplification.} We examine the influence of prototype simplification on performance and subsequence lengths. We use the previous settings with $\lambda_d=0.01$ on the UniProtKB and Yelp Reviews. There is no significant difference in accuracy on both datasets. However, with simplification applied, the average prototype (sub)sequence lengths are decreased from 20.1 to 15.1 on Yelp Reviews and 274.5 to 130.7 on UniProtKB dataset. \looseness=-1

\subsection{Human Evaluation of Interpretability}

The interpretability of a machine learning model is a subjective concept, which is often regarded to be difficult to evaluate computationally \cite{Lipton18mythos,doshi2017towards}. Thus, we conduct a quantitative evaluation of the interpretability of \systemname~through experiments with human subjects. With the prototype learning structure, we aim to answer the following questions: 1) How understandable and accurate are the prototypes in explaining the predictions of the input sequences? 2) How would the incorporation of human knowledge (Section~\ref{sec:method-refine}) influence the performance and interpretability of \systemname? We use the \systemname~trained on Yelp Reviews for binary sentiment classification  (Section~\ref{sec:exp-yelp}) for evaluation. The model has 80 effective prototypes (\ie, the associated weight $\max(\mathbi{w}_i) > 0.1 \max(\mathbi{W})$).  \looseness=-1

\begin{figure}[ht]
\includegraphics[width=\linewidth]{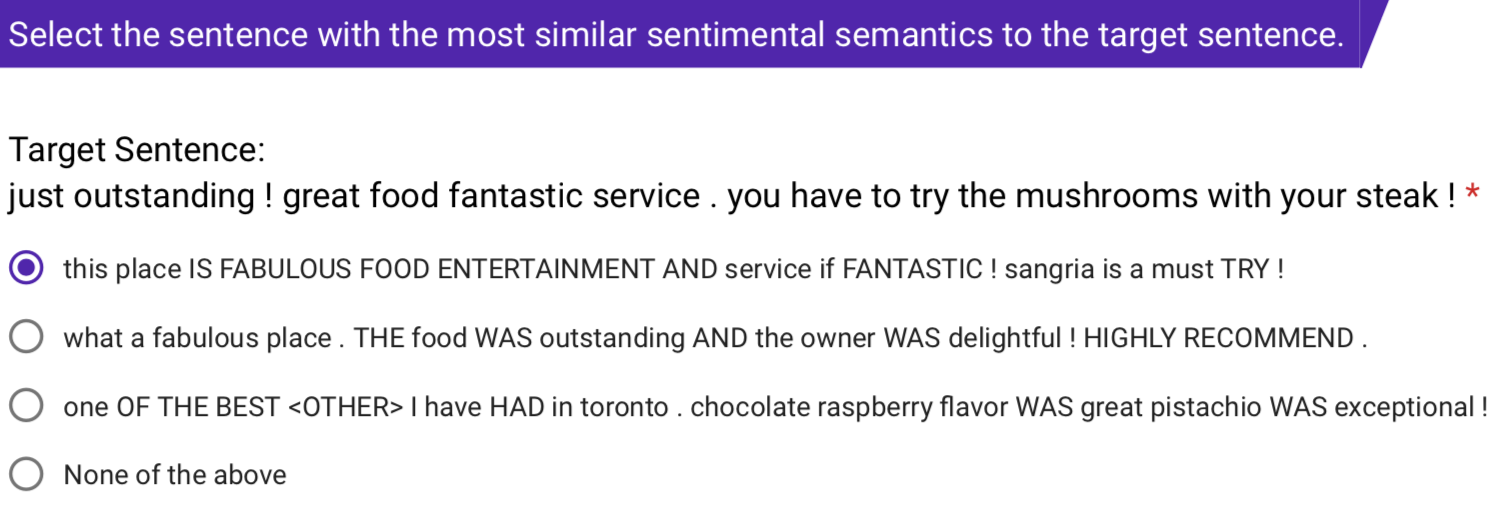}
\vspace{-0.25in}
\caption{Example question of the user experiment. The uppercase words represent simplified prototype sequences.}
\label{fig:yelp-user-study}
\vspace{-0.15in}
\end{figure}

\textbf{Experiment Setup}. To evaluate the interpretability of the explanations, we recruit human participants on Amazon Mechanical Turk, who are non-experts in machine learning. Directly asking whether an explanation is interpretable or accurate is very subjective and varies for different people. Thus, we adopt a relative measure by asking the participants to select one of three prototype sentences that expresses the most similar sentiment to a given input sentence. The prototype in the model that has the largest similarity score to the input sentence is regarded as the proposed answer by the model and is presented as one of the options. The other options are randomly selected from the rest of the prototypes. We also include a ``None of the above'' as the fourth option. The input sentences are selected from the validation set with stratified sampling. That is, we divide the sequences into groups according to their most similar prototypes, and use the groups as the strata for sampling. An example question is shown in \autoref{fig:yelp-user-study}.  \looseness=-1

\begin{table}[bth]
  \caption{Average accuracy of human subjects and \systemname~ before and after updating the model via interaction. The most voted option by human subjects is used as the correct answer. The accuracy is calculated over questions that are \textbf{not} most voted as ``None of the above''.}
  \vspace{-0.05in}
  \label{tab:yelp-mturk}
	\centering%
    \begin{tabular} {c c c c}
    \hline
    & Model acc. & Human acc. & None of the above \\
    \hline
    Before & 0.618 & 0.667 (SD: 0.113) & 0.131  \\
    After & \textbf{0.682} & 0.698 (SD: 0.142) & 0.131 \\
    \hline
    \end{tabular}
\end{table}

We sample 70 questions and split them into four questionnaires. We gather 20 responses from different human subjects for each of the questionnaires. After filtering the responses that failed quality check (\eg, consistency check of the answers of duplicate questions), each question has 12.5 valid responses on average. We use the most voted options by human subjects as the correct answer of each question and computes the accuracy of human and the model. The result is summarized in the first row in \autoref{tab:yelp-mturk}.  \looseness=-1

\textbf{Interacting with sequence prototypes.} To study how the input of human knowledge would affect the interpretability of the model, we use the feedback from the user study as a source of human knowledge to update the model (as described in Section~\ref{sec:method-refine}) and then run a second round of experiment on MTurk. Based on the result of the first round user experiment, we update the model to improve the quality of the prototypes. The update protocol is as follows. For each of the wrongly answered question, we check the prototype sequence that is proposed as the answer by the model, as well as its neighboring sequences in the validation set. If the neighboring sequences do not have consistent sentiment (with subjective judgment), we would delete this prototype. If the neighboring sequences do have consistent sentiment, but the provided prototype is not representative enough (\eg, part of the sentence has misleading meaning), a new sentence is selected from the neighboring sentences to replace the old prototype.  \looseness=-1

Following the above protocol, we updated 13 prototypes and removed 5 prototypes. After the incremental training completes, the performance of the model on the test set is basically unchanged (slightly increased by 0.1\%). Then we run the second user experiments with the same procedure. An average of 12.3 valid responses is collected for each question. 

\textbf{Result.} As shown in the second row in \autoref{tab:yelp-mturk}, the accuracy of the model's proposed answers increased significantly from 61.8\% to \textbf{68.2\%}, which is only 1.6\% lower than human accuracy. The result of the first experiment indicates that there is still a gap between the quality of the model's generated explanations and the human standard. However, the second experiment shows that the incorporation of human knowledge via our proposed interaction scheme could be very helpful in improving the interpretability of the model.  \looseness=-1

%
%

\section{Conclusion and Future Work}

In this paper, we presented an interpretable and steerable deep sequence modeling technique called \systemname. The technique combines prototype learning and RNNs to achieve both interpretability and high accuracy. Experiments and case studies on four different real-world sequence prediction/classification tasks demonstrated that \systemname~ is not only as accurate as other state-of-the-art machine learning techniques but also much more interpretable. In addition, large scale user study on Amazon Mechanical Turk demonstrated that for familiar domains like sentiment analysis on texts, \systemname~ is able to select high quality prototypes that are well-aligned with human knowledge for prediction and interpretation. Furthermore, \systemname~ obtained better interpretability without loss of performance by incorporating the feedback from the user study to update the prototypes, demonstrating the benefits of involving human-in-the-loop for interpretable machine learning. Future works include applying the technique to other sequence data and developing interactive user interface for updating the prototypes. \looseness=-1

\begin{acks}
This research was partially supported by Hong Kong TRS grant T41-709/17N.

\end{acks}

\bibliographystyle{ACM-Reference-Format}
\balance
\bibliography{bibliography}

%
%

\newpage
\appendix

\section{Prototype Simplification via Beam Search}

The beam search algorithm that we used for prototype simplification is shown in \autoref{algo:beam-search}. The $\textsc{BestCandidates}(\mathcal{S}, w)$ takes a set of sequences $\mathcal{S}$, computes the score using \autoref{eq:projection-simplification} for each sequence, and returns $w$ sequences with the lowest scores. The algorithm terminates when the subsequences are not reducible, or there is no better remove-one subsequences than the existing sequences in the set of candidates $\mathcal{S}$.

\begin{algorithm}[h]
\SetKwInput{Input}{Input}
\SetKwInput{Parameters}{Parameters}
\SetKwInput{Output}{Output}
\Input{encoder $r$, training data $\mathcal{X}$, prototype $\mathbi{p}_i$} 
\Parameters{beam width $w$}
\Output{projected prototype $\hat{\mathbi{p}}_i$, subsequence $\mathbi{s}$}
\tcc{Find $w$ sequences with minimum distance to $\mathbi{p}_i$}
$\mathcal{S} \gets \textsc{BestCandidates}(\mathcal{X}, w)$\;
$\mathbi{s}_{opt} \gets NULL$\;
\While{$\mathcal{S} \neq \emptyset$} {
    $\hat{\mathcal{S}} \gets \emptyset$\;    
    \For{$\mathbi{s} \in \mathcal{S}$}{
        $\mathbi{s}_{opt} \gets \textsc{BestCandidates}(\{\mathbi{s}_{opt},\mathbi{s} \}, 1)$\;
        \uIf{$|\mathbi{s}| \geq 1$}{
            \tcc{remove-one sub-sequences}
            $\hat{\mathcal{S}} \gets \hat{\mathcal{S}} \cup \{\mathbi{s}/e_i \mid e_i \in \mathbi{s}\}$\;
        }
    }
    \tcc{Find $w$ subsequences with the lowest scores using \autoref{eq:projection-simplification}, return $\hat{\mathcal{S}}$ if there are less than $w$ sequences in it.}
    $\mathcal{S} \gets \textsc{BestCandidates}(\hat{\mathcal{S}}, w)$\;
} 
$\mathbi{p}_i \gets r(\mathbi{s}_{opt})$\;
 \caption{Beam Search}
 \label{algo:beam-search}
\end{algorithm}





\section{Experiment Details}
\subsection{Data Processing of Yelp Reviews}

For sentiment classification tasks on Yelp Reviews, we first filtered the reviews to contain only ``Restuarant'' reviews according to ``category'' field of the business being reviewed. We then tokenize the review texts into sequences of words using NLTK. For human evaluation purpose, we filtered reviews with length (number of words) over 25. For both binary classification and fine-grained classification, we balanced the classes by down-sampling. The size of the largest class is no more than twice the size of the smallest class. The vocabulary sizes are 6287 and 6792. We using word embedding of size 100 for all the models. The embedding are jointly trained with the models.

\subsection{Post-processing of Human Evaluation Data}

We partition the 70 questions evenly to 4 questionnaires (each with 17 and 18 questions) to prevent participants becoming overwhelming. We also add three additional quality check questions (\eg, duplicate questions with options in a different order, or questions with obvious correct answer). 

We filtered the responses that fails more than 1 quality-check questions. Then we compute the correct answer of each question using the most voted option. We further filter responses that have have accuracy lower than 50\%. Then we finally compute the human accuracy and the model accuracy using the most voted options as the correct answers.

\section {Supplement Examples}

\subsection{Sentiment Classification on Yelp Reviews}

\begin{figure}[hbt]
\includegraphics[width=\linewidth]{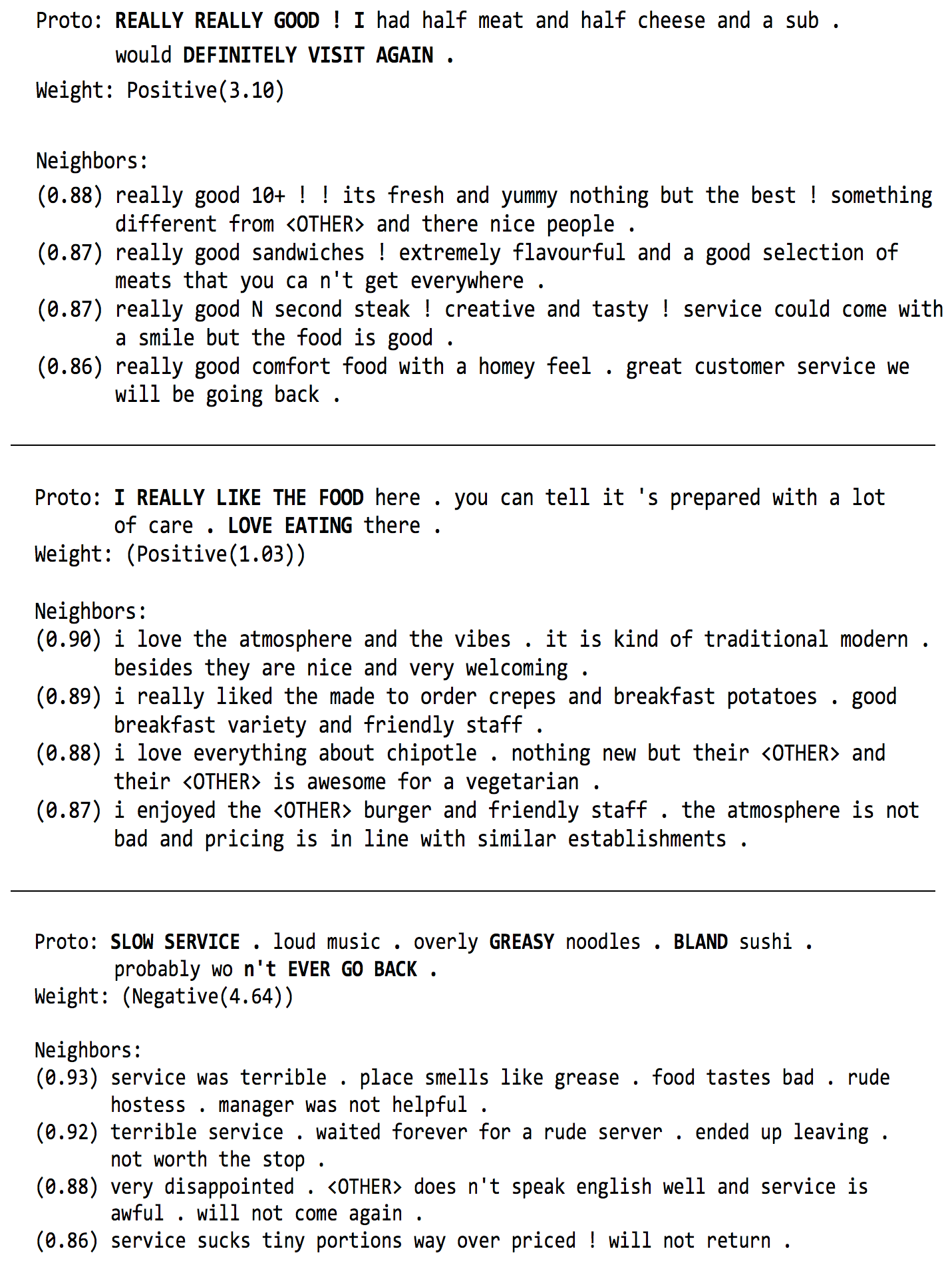}
\caption{Additional examples of learned prototype sequences on Yelp Reviews. The bold uppercase texts are the simplified subsequences.}
\label{fig:supp-yelp}
\end{figure}

\newpage

\subsection{ECG Heartbeat Classification}

\begin{figure}[bth]
\includegraphics[width=\linewidth]{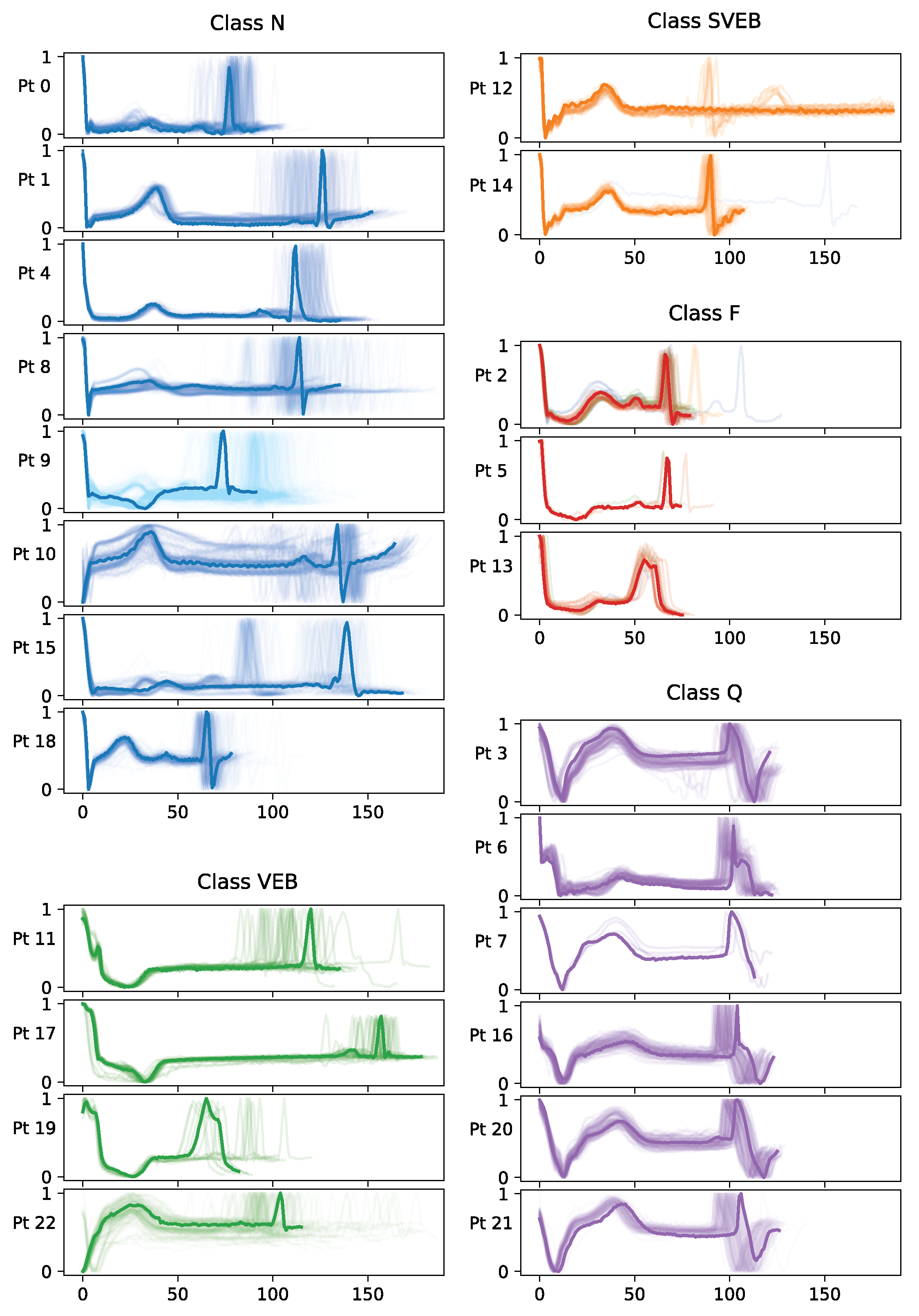}
\caption{Complete list of time series prototypes learned form the ECG heartbeats. Bold lines represent the prototype signals, and transparent lines shows are test signals close to the prototypes.}
\label{fig:ecg-prototype-complete}
\end{figure}



\end{document}